\documentclass[final,1p,times]{elsarticle}
\usepackage{amssymb}
\usepackage{amsmath}
\usepackage{acronym}
\usepackage{algorithm}
\usepackage{algorithmic}
\usepackage{makecell}
\usepackage[inline]{enumitem}
\usepackage{bbm}
\usepackage[skip=2pt]{caption}
\usepackage{natbib}
\usepackage{xcolor}
\usepackage{booktabs}
\usepackage{hyperref}
\usepackage{rotating}
\newcommand{\blue}[1]{#1}

\acrodef{TCN}{temporal convolutional network}
\acrodef{CNN}{convolutional neural network}
\acrodef{RNN}{recurrent neural network}

\journal{International Journal of Forecasting}

\begin{document}

\begin{frontmatter}

\title{Parameter Efficient Deep Probabilistic Forecasting}

\author[auth1]{Olivier Sprangers}
\ead{o.r.sprangers@uva.nl}
\author[auth2]{Sebastian Schelter}
\ead{s.schelter@uva.nl}
\author[auth3]{Maarten de Rijke}
\ead{derijke@uva.nl}

\address[auth1]{AIRLab, University of Amsterdam}
\address[auth2, auth3]{University of Amsterdam}

\begin{abstract}
Probabilistic time series forecasting is crucial in many application domains such as retail, ecommerce, finance, or biology. 
With the increasing availability of large volumes of data, a number of neural architectures have been proposed for this problem. 
In particular, Transformer-based methods achieve state-of-the-art performance on real-world benchmarks. 
However, these methods require a large number of parameters to be learned, which imposes high memory requirements on the computational resources for training such models.

To address this problem, we introduce a novel \textit{Bidirectional Temporal Convolutional Network} (BiTCN), which requires an order of magnitude less parameters than a common Transformer-based approach. 
Our model combines two Temporal Convolutional Networks (TCNs): the first network encodes future covariates of the time series, whereas the second network encodes past observations and covariates. 
We jointly estimate the parameters of an output distribution via these two networks. 

Experiments on four real-world datasets show that our method performs on par with four state-of-the-art probabilistic forecasting methods, including a Transformer-based approach and WaveNet, on two point metrics (sMAPE, NRMSE) as well as on a set of range metrics (quantile loss percentiles) in the majority of cases. Secondly, we demonstrate that our method requires significantly less parameters than Transformer-based methods, which means the model can be trained faster with significantly lower memory requirements, which as a consequence reduces the infrastructure cost for deploying these models.
\end{abstract}

\begin{keyword}
Probabilistic forecasting, Temporal convolutional network, Efficiency in forecasting methods
\end{keyword}

\end{frontmatter}

\section{Introduction}

Time series forecasting is crucial to many application domains such as retail, ecommerce, finance or biology. The classical approach to forecasting is to learn a single model for each time series separately. However, creating such models often requires careful manual intervention from forecasting practitioners, which becomes impractical when the number of time series to be forecast is large (e.g., when the task is to forecast store demand for each product of a retail chain~\cite{bose_probabilistic_2017, taylor_forecasting_2018}). Moreover, decision makers often prefer probabilistic forecasts to point forecasts because they are interested in a quantification of the uncertainty of the forecast. However, probabilistic forecasts may require a set of models for each time series, which further increases the disadvantages of using a single model for each time series. In this work, we address this problem of probabilistic forecasting for multiple time series.

\blue{Recently, a number of neural network architectures have been proposed \cite{fischer_deep_2018,laptev_time-series_2017,li_diffusion_2018} to address the disadvantages of classical forecasting methods. These neural architectures leverage the increasing availability of data in the aforementioned application domains, and enable the training of a single model that can produce forecasts for multiple time series over multiple time steps. The Transformer~\cite{vaswani_attention_2017} is one such neural architecture that has recently shown state-of-the-art performance on a set of real world forecasting datasets~\cite{li_enhancing_2019}. However, it requires a large number of parameters to achieve these state-of-the-art results compared to existing neural architecture-based probabilistic forecasting methods. }

In this work, we demonstrate that it is possible to achieve on-par or better results than a Transformer with an architecture that requires an order of magnitude less parameters. We achieve this by 
\begin{enumerate}[label=(\arabic*)]
\item Smartly encoding available future information; 
\item Applying a simple temporal convolutional architecture; and 
\item \blue{Leveraging the Student's t(3)-distribution as a loss function to optimize parameters of our method. }
\end{enumerate}
\blue{Our \textit{Bidirectional Temporal Convolutional Network} (BiTCN) is based on two Temporal Convolutional Networks (TCNs), as detailed in Section~\ref{sec:model}. The first network encodes future covariates of the time series, whereas the second network encodes past observations and covariates. This method allows us to preserve temporal information of sequence data, and is computationally more efficient by requiring fewer sequential operations than the commonly used bidirectional LSTM. The output of both our networks is used to estimate the parameters of a Student's t(3)-distribution of our forecast. }

\blue{Next to introducing BiTCN, the contributions of this paper are threefold: 
\begin{itemize} 
\item We empirically verify that we achieve state-of-the-art forecasting performance with BiTCN on four real-world datasets (Sections~\ref{sec:exp-setup} and~\ref{sec:results}). We evaluate on point forecast error metrics (sMAPE, NRMSE) as well as on probabilistic forecast error metrics (quantile loss percentiles).
\item We show how BiTCN is designed \begin{enumerate*}[label=(\arabic*)] \item to require an order of magnitude less parameters than the second-best scoring Transformer-based method, and \item to employ a simpler architecture than WaveNet~\citep{oord_wavenet_2016} -- an autoregressive neural network based on a TCN -- which placed second in a Kaggle forecasting competition~\citep{kechyn_sales_2018}. \end{enumerate*} The result is that our model can be trained using cheaper hardware because it requires less memory (approx.\ 2--4x less GPU memory, depending on batch size). Moreover, our method also requires at least 20\% less energy during training, thereby reducing cost to train the model. 
\item We demonstrate the benefits of choosing a Student's t(3)-distribution for the probabilistic forecasting setting~(Section~\ref{subsec:studentteffect}), which enables us to eliminate a training hyperparameter compared to using a Gaussian distribution for probabilistic forecasting, and it results in a more stable training regime.
\end{itemize}}

\section{Related Work}
\label{sec:related}

Time series forecasting is a broad topic that is being studied in various scientific disciplines, such as econometrics, economics and machine learning. \blue{Traditional forecasting models such as ARIMA \cite{box_distribution_1970} and Exponential Smoothing \cite{hyndman_forecasting_2008} rely on considering the time series individually, and thus create separate models for each time series (typically referred to as local methods \cite{montero-manso_principles_2021}). These local methods may be more difficult to apply in contemporary large-scale forecasting applications as maintaining individual models per time series may be impractical and these methods typically struggle to forecast unseen time series with little or no past observations. Moreover, \cite{montero-manso_principles_2021} have also demonstrated such local methods are typically outperformed by global methods (i.e. methods that create a single global model across all time series).}

Recently, neural networks (in the form of sequence-to-sequence models) have been successfully applied to various autoregressive problems, such as speech generation with WaveNet~\citep{oord_wavenet_2016} and probabilistic forecasting with DeepAR~\citep{salinas_deepar_2019}. We refer to \citep{mariet_foundations_2019} for a more formal introduction to sequence-to-sequence modeling for time series, \blue{and to \cite{montero-manso_principles_2021} for a study of the benefits of global models compared to local models for time series}.

\paragraph{Recurrent neural networks} RNNs, usually in the form of Long-Short Term Memory (LSTM) \cite{hochreiter_long_1997} modules, have been widely applied to the forecasting problem; \cite{fischer_deep_2018,laptev_time-series_2017,li_diffusion_2018} provide recent examples of applications to financial markets, taxi services and traffic forecasting,  respectively. Moreover, RNN models have been succesful at winning several forecasting competitions, such as the LSTM-based ES-RNN method that won the M4 forecasting competition \cite{makridakis_m4_2020}, and the LSTM-based method that won the Kaggle Webtraffic competition.\footnote{\url{https://github.com/Arturus/kaggle-web-traffic}}
The downside of employing RNN models in forecasting is that the recurrent nature of the model leads to slow training times, and long-term dependencies may not be properly captured. 
To improve the long-term memory of RNNs, attention mechanisms \cite{bahdanau_neural_2014} have been combined with RNNs in \cite{lai_modeling_2018,chen_tada_2018}. However, these methods still rely on the sequential training of RNN modules. \blue{\cite{hewamalage_recurrent_2021} provide a recent extensive study comparing RNN-based forecasting methods to classical approaches such as ARIMA and ETS, which concludes that RNN-based forecasting methods can be competitive to classical approaches in many scenarios.}

\paragraph{Attention-only models} The Transformer \cite{vaswani_attention_2017} has been introduced in the domain of Natural Language Processing (NLP) to enable efficient parallel training of sequence-to-sequence problems. Recently, the Transformer has been adopted to the problem of probabilistic forecasting in \cite{li_enhancing_2019}, who employ a decoder-only model with convolutional sparse attention to reduce memory consumption of the base Transformer and enhance its forecasting performance. 
This method currently achieves state-of-the-art results on the various public datasets commonly used in probabilistic forecasting papers. 

\paragraph{Temporal convolutional networks} TCNs employ stacked dilated Convolutional Neural Networks (CNNs) to overcome the limitations of RNN \cite{bai_empirical_2018}. The benefit of this architecture is that it allows for fast training like the Transformer model, whilst having a significantly lower memory consumption. This enables larger batch sizes during training which, in turn, speeds up training. Temporal convolutional networks have been applied to autoregressive problems in speech generation with Wavenet~\citep{oord_wavenet_2016}, and achieved the second place in a Kaggle sales forecasting competition~\citep{kechyn_sales_2018}. A more recent method closely related to ours is introduced in \cite{sen_think_2019}, where a matrix factorization method is combined with a standard TCN to provide point forecasts. 

Our method is different in that we do not employ matrix factorization, and study the problem of probabilistic forecasting instead of point forecasts. Probabilistic forecasting with TCN has recently been conducted by \cite{chen_probabilistic_2019}. Our work differs from this work in that \begin{enumerate*}[label=(\arabic*)] \item our core temporal module is simpler as it uses fewer components, \item our focus is on parameter efficient probabilistic forecasting, and \item we introduce a forward-looking module to encode future covariates. \end{enumerate*}

\section{Methodology}
\label{sec:model}
We introduce the problem setting of probabilistic forecasting, and describe the core components of our method. We end the section by detailing the choice of our loss function.

\paragraph{Probabilistic forecasting}\label{par:probforecasting}
A time series is a sequence of ordered measurements \( \{ {y}_{t}, {y}_{t+1}, \dots \}\), in which we assume the timestep \(t\) to be constant (e.g., a day, an hour). 
Denote the set of \(N\) time series as \( \{ \vec{y}_{i,1:t_0} \}^N_{i=1} \) and \( \{ \vec{a}_{i,1:t_0} \}^N_{i=1} \) a set of additional attributes. 
We are interested in modeling the conditional distribution 
\begin{equation}
\begin{split}
p(\vec{y}_{i, t_0:T} \vert \vec{y}_{i, 1:t_0}, &\vec{a}_{i, 1:T} ; \mu_\theta, \sigma_\theta) \\
  = &\prod_{t=t_0}^{T} p\left(\vec{y}_{i, t} \vert \vec{y}_{i, 1:t-1}, \vec{a}_{i, 1:T} ; \mu_\theta, \sigma_\theta \right)  \\
  = &\prod_{t=t_0}^{T} p\left(\vec{y}_{i, t} \vert \vec{x}_{i, 1:T} ; \mu_\theta, \sigma_\theta \right),
\end{split}
\end{equation}
where \(t_0\) and \(T\) denote the start and end of the forecast,  respectively, and \((\mu_\theta, \sigma_\theta)\) the location and scale parameters of a distribution parameterized by \(\theta\), which we learn with our model. 
We estimate separate output distribution parameters for each timestep in the forecast window. The input to our network consists of the concatenation of lagged target variables \(\vec{y}_{lag}\) and additional attributes in the form of numerical covariates \(\vec{a}_{cov}\) (e.g., a day-of-the-week indicator) and categorical covariates \(\vec{a}_{cat}\) (e.g., a time series identifier). We denote this concatenation by \(\vec{x}\).

\paragraph{Dilated convolutions} We observe that future covariates on time series are usually available at the time of forecasting, such as item 
 information, or time indicators (e.g., day-of-the-week or indicators for promotion days or holidays). We would like to encode all such knowledge of the future into a latent state on which the forecast of the current timestamp can be conditioned. To achieve this, we create a TCN consisting of dilated convolutions that look \textit{forward} in time, instead of backward. More formally, the \textit{forward} dilated convolution operation \(F\) on an element \(s\) of a sequential input \(\mathbf{x} \in \mathbb{R}^n\) and a filter \(f \in \mathbb{R}^k\) can be defined as \cite{bai_empirical_2018}:
\begin{align}\label{eq:dilcausconv}
	F(s) = \sum_{j=0}^{k-1}f(j) \cdot \mathbf{x}_{s + d\cdot j}, 
\end{align}
with filter size \(k\), dilation factor \(d\), and \(s + d \cdot j\) indicating the forward steps. For a backward (causal) dilated convolution, \(s + d \cdot j\) in \eqref{eq:dilcausconv} becomes \(s - d \cdot j\). 
\blue{We stack \(N\) layers with dilation \(2^{i -1}\) for layer \(i\) to obtain a `receptive field` for the TCN of size \(1 + (k-1)(2^N-1)\). The receptive field is the effective sequence length the network can condition its forecast on. For example, if a forecasting problem requires taking into account sequences of a length up to 500 time steps, example valid options for the kernel size and number of layers N would be \((k = 3, N = 8)\),  \((k = 7, N=7 )\) or \((k = 11 , N=6 )\)}. It is necessary to add right (left) padding of size \( (k-1)\cdot 2^{i-1}\) for the forward (backward) convolution at each \(i^{th}\) layer to maintain a constant sequence length throughout the network. Many existing forecasting methods already incorporate future covariate information. However, our approach is novel due to the fact that: 
\begin{enumerate}[label=(\arabic*)]
\item The temporal structure of the covariates is maintained vis-a-vis feed-forward networks to incorporate such information, and 
\item Dilated convolutions enable more efficient training than, e.g., a bidirectional LSTM, which requires more sequential operations during training (we refer to Section~\ref{sec:results} for a discussion on the computational efficiency). 
\end{enumerate}
We considered an alternative design with an unmasked multi-head attention module \cite{vaswani_attention_2017} as a `look-forward' layer. We decided against this design however, as preliminary experiments indicated relatively poor results in terms of parameter / performance balance (many additional parameters required for incremental accuracy gains) -- see Section~\ref{sec:results} for a discussion on the complexity of multi-head attention layers.

\begin{figure}[t]
\centering
\centerline{\includegraphics[width=\columnwidth]{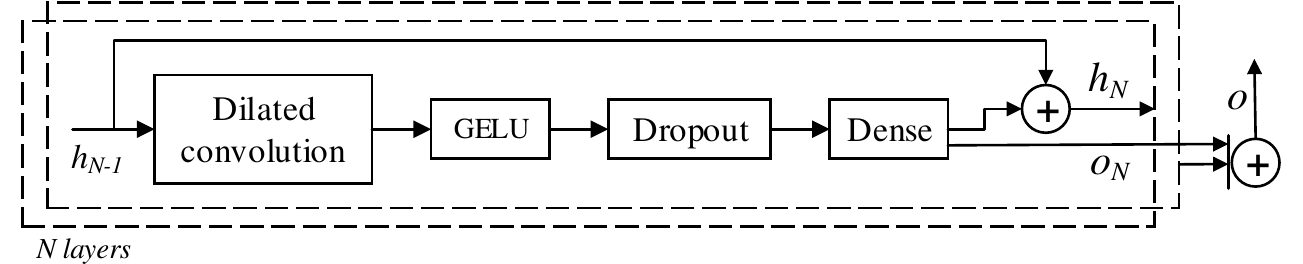}}
\caption{Temporal block}
\label{fig:temporal_block}
\end{figure}

\paragraph{Temporal blocks} Inspired by the work of \cite{bai_empirical_2018,oord_wavenet_2016}, we construct temporal blocks by stacking dilated convolutions. A single layer of our TCN is displayed in Figure~\ref{fig:temporal_block}. Each TCN layer in our network consists of a block that contains a dilated convolution, a GELU activation \protect\cite{hendrycks_gaussian_2018}, dropout and a dense layer. \blue{A Gaussian Error Linear Unit (GELU) multiplies an input \(x\) with the cumulative distribution function \(\Phi(x)\) of the Gaussian distribution, i.e.
\begin{align}
GELU(x) &= x P(X \le x)  = x \Phi(x) \\
				&\approx 0.5 x \left( 1 + \tanh \left[ \sqrt{2 / \pi} \left( x + 0.044715x^3 \right) \right] \right)  \nonumber
\end{align}
and provides a `softer' activation than the commonly used Rectified Linear Unit (ReLU), which  generally improves performance \cite{hendrycks_gaussian_2018}.}

Each layer produces a hidden state \(h_N\) and an output \(o_N\); the latter is summed over all layers to provide the output of the network.  
Weight normalization \protect\cite{salimans_weight_2016} is applied to the dilated convolution and dense modules. 
Our temporal block is simpler than the ones used in WaveNet, as our temporal block does not require a \textit{gated activation unit}, but instead relies solely on the GELU activation as non-linearity. 
We will show in Section~\ref{subsec:fperformance} that this simplification has a positive effect on performance.

\begin{figure}[t]
\centering
\centerline{\includegraphics[width=\columnwidth]{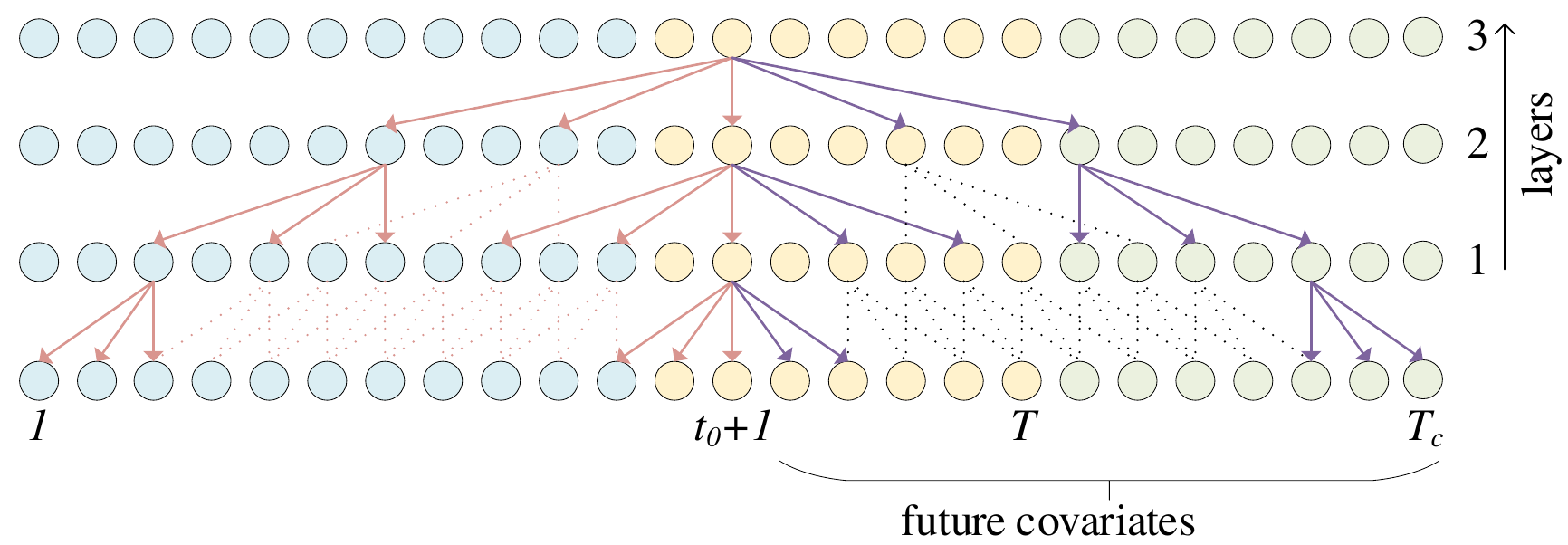}}
\caption{An illustration of how 3 stacked TCN layers enable conditioning the forecast at \(t = t_0 + 1\) on both past and future information using forward and backward dilated convolutions with kernel size 3 and dilation \(2^{i-1}\) for the \(i\)-th layer. The blue dots represent the input sequence, the yellow dots the output sequence and the green dots the additional future covariates on which the forecast can be conditioned. The red connections indicate the \textit{backward} looking convolutions, and the purple connections the \textit{forward} looking convolutions. For clarity purposes, some inner convolutional connections are shown with dashed lines.}
\label{fig:dilatedconvs}
\end{figure}

Our method employs two temporal blocks; the first block consists of \(N\) layers and encodes the past observations and covariates of time series with backward dilated convolutions, whereas the second block consists of \(N+1\) layers and encodes future covariates with forward dilated convolutions. 
\blue{The additional layer in the second block is required to enlarge the receptive field of the forward looking module, as the sequence length of the covariates and categorical inputs exceeds the dimension of the input and output sequence length in order to facilitate sufficient look-forward for later timesteps in the forecast. }
This is to ensure that later timesteps of the forecast also receive sufficient future covariate information to condition the forecast on. 
A 3-layer example of how these temporal blocks employ backward and forward dilated convolutions to condition the forecast is pictured in Figure~\ref{fig:dilatedconvs}.

\begin{figure}[t]
\centering
\centerline{\includegraphics[width=\columnwidth]{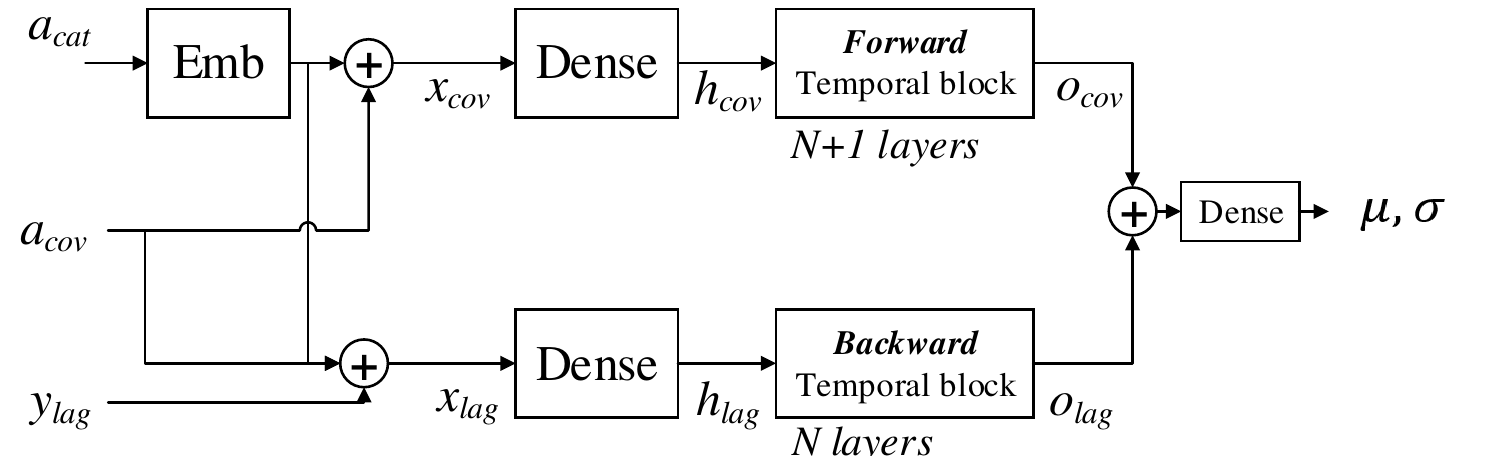}}
\caption{Overall network of BiTCN. First, in the top half, the categorical inputs \(a_{cat}\) are embedded using an embedding layer. The result is combined with the available covariates \(a_{cov}\) to result in the input \(X_{cov}\), which includes all available (covariate) information about the past and future. This is then led through a linear layer and a temporal block that looks forward across all the future information. In the bottom half, the lagged target \(y_{lag}\) is combined with the available historical covariate and categorical information to result in the input \(X_{lag}\). This is then led through a linear layer and an autoregressive temporal block (i.e. standard dilated convolutions). The outputs of both halfs are joined by adding them together at the right time stamp and a final linear layer provides the location and scale of our output distribution. }
\label{fig:network}
\end{figure}

Within the forward TCN, grouped convolutions are employed to reduce the amount of parameters needed. 
The intuition behind this choice is that the future covariates usually contain less information than the past covariates and lags, and thereby require less parameters for encoding. 
We employ dense layers with dropout before the temporal blocks to scale the input dimensions to the hidden dimension of the network. 
Finally, dense output layers provide the location \(\mu\) and scale \(\sigma\) of the output distribution. 
\blue{These layers are activated with a softplus activation to ensure that their output remains positive (this activation may be removed in case the possibility of a negative output is desired) and a small number \(\epsilon\) is added to ensure numerical stability of the scale output.} The pseudocode of our resulting method, which we call \textit{Bidirectional Temporal Convolutional Network} (BiTCN), is shown in Algorithm~\ref{alg:bitcn} and graphically depicted in Figure~\ref{fig:network}. 

\begin{algorithm}
\caption{BiTCN pseudocode}
\label{alg:bitcn}
\begin{flushleft}
\textbf{Input}: \(\vec{y}_{lag} \in \mathbb{R}^{T \times d_{batch} \times d_{input}}\), \(\vec{a}_{cov} \in \mathbb{R}^{T_{c} \times d_{batch} \times d_{cov}}\), \(\vec{a}_{cat} \in \mathbb{R}^{T_c \times d_{batch} \times d_{cat}}\) \\
\textbf{Output}: \(\vec{\mu}, \vec{\sigma}\) 
\end{flushleft}
\begin{algorithmic}[1]
\STATE \(\vec{a}_{emb} = \texttt{Embedding}(\vec{a}_{cat})\)
\STATE \(\vec{x}_{lag} = \texttt{Concat}(\vec{y}_{lag}, \vec{a}_{cov}\text{[:\(d_l\)]}, \vec{a}_{emb}\text{[:\(d_l\)]})\)
\STATE \(\vec{x}_{cov} = \texttt{Concat}(\vec{a}_{cov}, \vec{a}_{emb})\)
\STATE \(\vec{h}_{lag} =\texttt{Drop}(\texttt{Dense}(\vec{x}_{lag})) \)
\STATE \(\vec{h}_{cov} =\texttt{Drop}(\texttt{Dense}(\vec{x}_{cov})) \)
\STATE \(\vec{o}_{lag} = \texttt{Backward Temporal Block}(\vec{h}_{lag}) \)
\STATE \(\vec{o}_{cov} = \texttt{Forward Temporal Block}(\vec{h}_{cov}) \)
\STATE \(\vec{o} =  \texttt{Concat}(\vec{o}_{cov}\text{[:\(d_l\)]}, \vec{o}_{lag})\)
\STATE \(\vec{\mu} = \texttt{SoftPlus}(\texttt{Dense}(\vec{o})) \)
\STATE \(\vec{\sigma} = \texttt{SoftPlus}(\texttt{Dense}(\vec{o})) + \epsilon \)
\STATE \textbf{return} \(\vec{\mu}, \vec{\sigma}\)
\end{algorithmic}
\end{algorithm}

\paragraph{Output distribution}
\blue{Although BiTCN allows the use of many parametric (e.g., Gaussian) or non-parametric (e.g., Quantile) distribution functions, we use a Student's t-distribution with three degrees of freedom in our experiments as fixed parameterized distribution. 
The probability density function of this distribution for a variable \(\vec{y}\) is given by:
\begin{align}
	f(\vec{y}) = \frac{2}{\pi \sqrt{3}\left(1 + \frac{\vec{y}^2}{3}\right)^2}
\end{align}}
and we aim to minimize the log-likelihood loss
\begin{align}
  \mathcal{L}(\theta \vert \vec{y}) = \log \left(f\left(\frac{\vec{y} - \mu_\theta}{\sigma_\theta}\right) \right),
\end{align}
where \(\mu\) and \(\sigma\) are the outputs of our network for a given time series at a certain timestep. There are several benefits of using this distribution. First, the Student's t-distribution can be seen as a fat-tailed version of the Gaussian distribution, which makes it a natural candidate for estimating real-valued quantities without prior information on the true distribution. The benefit of a fat-tailed distribution is two-fold: 
\begin{enumerate}[label=(\arabic*)]
\item This allows our model to capture rare events better, and 
\item The log-likelihood loss is numerically stable around a large range of input values.
\end{enumerate}
The disadvantage of fixing a distribution a priori is the imposition of a defined distribution on the output, which may not reflect the true underlying distribution properly. However, we observe this to be no issue in practice. 

To facilitate a better comparison between architectures, our loss function is used for each method in our experimental section comparing forecasting accuracy across competing methods. 
We will demonstrate the benefits of choosing the Student's t(3)-distribution over the Gaussian distribution in Section~\ref{subsec:studentteffect}.

\section{Experimental Setup} \label{sec:exp-setup}

\paragraph{Datasets} We employ four real-world datasets to investigate the probabilistic forecasting performance of the BiTCN architecture. 
\begin{itemize}
	\item \texttt{Electricity}\cite{noauthor_uci_nodate-3} Hourly electricity consumption for 370 clients over a 3 year period. The task is to forecast electricity consumption per client for the next 24 hours given the previous 7 days. The training/validation/test split is 80/10/10 based on date.
	\item \texttt{Traffic} \cite{noauthor_uci_nodate-4} 15 months of hourly data describing the occupancy rate, between 0 and 1, of different car lanes of the San Francisco bay area freeways. We forecast occupancy rate per lane for the next 24 hours given the previous 7 days. The training/validation/test split is 80/10/10 based on date.
	\item \texttt{Favorita}\cite{noauthor_corporacion_nodate} A dataset from Kaggle that contains 4 years of daily sales data for store-product combinations taken from a retail chain. The task is to forecast the log sales for all product-store combinations for the next 30 days based on the previous 90 days. The training data consists of data between 01-01-2013 and 02-09-2013. We validate on samples between the dates 03-09-2013 and 03-10-2013, and we test on the remaining data up to 31-03-2014. The target and lagged input variables are log-transformed.
	\item \texttt{WebTraffic}\cite{noauthor_web_nodate} 145k time series of daily Wikipedia pageviews in the period 2015-2017. The task is to forecast the number of pageviews for the next 30 days given the previous 90 days. We use a selection of 10k time series with the largest number of pageviews in the training period (which is up to 31-12-2016). 
\end{itemize}
We provide more detailed descriptions of the datasets in Table~\ref{tab:datasets} in~\ref{app:supplemental}.

The \texttt{Electricity} and \texttt{Traffic} datasets provide an indication of forecasting performance on relatively regular time series, with a small number of future covariates available to condition our forecast on (i.e., a few time-based features such as day of the week and hour of the day). 
\texttt{Favorita} and \texttt{WebTraffic} contain more irregular time series, and the first also contains a rich set of covariates. 
For each dataset, we add categorical covariates and numerical covariates consisting mainly of time series identifiers, time indicators (day-of-week, month) and other indicators (e.g., holiday). 
\blue{The time indicators are represented by two Fourier terms \cite{hyndman_forecasting_2018} to represent the periodic nature of, e.g., days or months (i.e., the first day of the week should be close to the last day of the week) as follows:}
\begin{align}
\text{covariate}_{\sin} &= \sin{\left(\text{covariate} \cdot \left(2 \cdot \frac{\pi}{\text{period}}\right)\right)} \\
\text{covariate}_{\cos} &= \cos{\left(\text{covariate} \cdot \left(2 \cdot \frac{\pi}{\text{period}}\right)\right)}.
\end{align}
The categorical and numerical covariates are assumed to be known a priori, i.e., we include these variables in the forward-look of BiTCN. 
For all datasets except \texttt{Favorita}, we employ the scaling mechanism from \cite{salinas_deepar_2019} for normalizing the outputs and lagged outputs to our network:
\begin{align}
\overline{\vec{y}} = \frac{\vec{y}}{1 + \frac{1}{t_0} \sum_{i=0}^{t_0} \vec{y}_i}.
\end{align}
Again, we refer to Table~\ref{tab:datasets} in the Appendix for further details on the datasets.

\paragraph{Baseline models}
We compare BiTCN against five state-of-the-art forecasting methods of different neural architectures:
\begin{itemize}
	\item \textit{DeepAR} \cite{salinas_deepar_2019}. An LSTM-based generic probabilistic forecasting framework. 
	\item \blue{\textit {ML-RNN} \cite{wen_multi-horizon_2018}. An encoder-decoder probabilistic forecasting framework that uses an LSTM to encode past observations and covariates and two MLP decoders to generate probabilistic forecasts. This method also uses future covariate information to condition its forecast on.}
	\item \textit{TransformerConv} \cite{li_enhancing_2019}. A Transformer-based forecasting model augmented with causal convolutions, which has shown state-of-the-art results on the \texttt{Electricity} and \texttt{Traffic} dataset.
	\item \textit{TCN} \cite{bai_empirical_2018}. A standard TCN, as employed for probabilistic forecasting by \cite{chen_probabilistic_2019}.
	\item \textit{WaveNet} \cite{oord_wavenet_2016}. which has been used to achieve the second place in a Kaggle sales forecasting competition \cite{kechyn_sales_2018}.
\end{itemize}

\blue{In addition, we compare BiTCN against three traditional forecasting methods and one popular machine learning package:
\begin{itemize}
	\item \textit{Seasonal Naive}. The seasonal naive baseline, where we simply take the observation from the last period as our forecast. Since our datasets exhibit clear seasonality (either daily or weekly), this provides a sensible baseline.
	\item \textit{ETS} \cite{holt_forecasting_2004}. Exponential smoothing method, including Holt-Winters' seasonal and trend components.
	\item \textit{Theta} \cite{assimakopoulos_theta_2000}. The theta method, a univariate forecasting method that uses the local curvature of the time series. 
	\item \textit{LightGBM} \cite{ke_lightgbm_2017}. A highly popular Gradient Boosting package on which the winning solution of the M5 forecasting competition \cite{makridakis_m5_2020} was based. 
\end{itemize}}

\paragraph{Training and optimization} 
We implemented, trained and evaluated BiTCN and each neural network method using PyTorch \cite{paszke_pytorch_2019} and our code can be found on Github.\footnote{\url{https://github.com/elephaint/pedpf}}
For DeepAR, we used the hyperparameters from \cite{salinas_deepar_2019}, for ML-RNN from \cite{wen_multi-horizon_2018} and for TransformerConv, we used the hyperparameters from \cite{li_enhancing_2019}. For TCN, WaveNet and BiTCN we selected a kernel size and hidden size to establish a comparable parameter budget as DeepAR, as this is the state-of-the-art in terms of parameter budget compared against in this study. An overview of the key model hyperparameters is given in Table~\ref{tab:experimentsetup}. Finally, we optimized learning rate and batch size for each dataset and method by performing a limited grid search using the following settings:
\begin{itemize}
	\item Learning rate: \(\{0.001, 0.0005, 0.0001\}\).
	\item Batch size: \(\{128, 256, 512\}\), except for DeepAR \blue{and ML-RNN} for which we used \(\{64, 128, 256\}\)
\end{itemize}

\noindent%
We run each experiment for 100 epochs with an early stopping criterion of 5 epochs. We train, validate and test each method for 5 different random seeds for the neural network weight initialization. We optimize the parameters of each method using Adam \cite{kingma_adam_2015}. The results of the limited grid search for each method and dataset are given in Figures~\ref{fig:hyperparam_electricity}--\ref{fig:hyperparam_webtraffic} in the Appendix. For the forecasting performance evaluation, we report the evaluation metrics on the test set of the best performing \{(learning rate, batch size)\} combination according to Figures~\ref{fig:hyperparam_electricity}--\ref{fig:hyperparam_webtraffic}. 

\blue{For the traditional forecasting methods, we use the \texttt{statsmodels} Python package and fit a model on each input sequence of our test set. For the probabilistic forecasts, we use the generated prediction intervals if the method provides these. For LightGBM, we create 9 separate models for each quantile \(0.1, 0.2,\dots,0.9\) using quantile regression and recursively apply each model to forecast each time step in our forecast. We use Optuna \cite{akiba_optuna_2019} to find the best hyperparameters for LightGBM, which we apply to all datasets.}

\begin{table*}[t]
\caption{Key model hyperparameters.}
\label{tab:experimentsetup}
\begin{center}
\blue{\begin{tabular}{l c c c c c c}
\toprule 
   & DeepAR & ML-RNN & TransformerConv & TCN & WaveNet  & BiTCN \\
\midrule
state size &40 &30 &\(d_{emb} + d_{cov} + d_{lag}\) & 20 & 20 & 12 \\ 
layers &3 &1 &3 &5 &5 &5  \\
kernel size &n.a. &n.a. & 9 &9 &9 &9 \\
heads &n.a. &n.a. &8 &n.a. &n.a. &n.a.  \\
dropout &0.1 &n.a. &0.1 &0.1 &n.a. &0.1  \\
receptive field &\(\infty\) &\(\infty\) &\(\infty\) & 249 &249 &497 \\
\bottomrule
\end{tabular}}
\end{center}
\end{table*}

\paragraph{Evaluation} 
We employ a set of point accuracy metrics and range accuracy metrics to evaluate forecasting performance. For point accuracy, we use symmetric Mean Absolute Percentage Error (sMAPE) and Normalized Root Mean Squared Error (NRMSE) \cite{alexandrov_gluonts_2020}:
\begin{align}
sMAPE &= \frac{1}{n}\sum_{t=t_0}^{n} \frac{2 \vert \vec{y}_t - \hat{\vec{y}}_t \vert}{\vert \vec{y}_t \vert + \vert \hat{\vec{y}}_t \vert} \\
NRMSE &= \frac{\sqrt{\frac{1}{n}\sum_{t=t_0}^{n} \left(\vec{y}_t - \hat{\vec{y}}_t\right)^2} \cdot \mathbbm{1}_{\vec{\overline{y}}_t \neq 0} }{\sum_{t=t_0}^{n}\vert \vec{y}_t \vert + \mathbbm{1}_{\vec{\overline{y}}_t = 0}},
\end{align}
where \(n = T - t_0\) denotes the number of forecast steps and \(\mathbbm{1}_{\vec{\overline{y}}_t = 0}\) is an indicator function to scale the metric when the observed target value equals zero. \blue{Note that the sMAPE ranges from \(0 - 200\%\). The NRMSE is essentially the normal root mean squared error, but normalized to account for differences in values of each individual time series. Each forecasting metric has its advantages and disadvantages; we also considered using a popular one-step ahead metric such as the MASE \cite{hyndman_another_2006}, but considered this inappropriate for our task (which is multistep forecasting).} For range accuracy, the normalized quantile loss function \cite{salinas_deepar_2019} is used for the quantiles \(p = \{0.1, 0.5, 0.9\}\):
\begin{align}
	Q(\vec{y}_i,\vec{\hat{y}}_i,p) &= 2 \cdot \vert (\vec{y} - \vec{\hat{y}}) \cdot (\mathbbm{1}_{\vec{y} \leq \vec{\hat{y}}} - p) \vert \\
	Q(\vec{y},\vec{\hat{y}},p) &= \frac{\sum_i Q(\vec{y}_i,\vec{\hat{y}}_i,p)}{\sum_i \vec{y}_i}.
\end{align}
Finally, we show the mean quantile performance over the 9 quantiles in the range \(p = \{0.1, 0.2, \dots, 0.9 \}\). 

\section{Results \& Discussion} \label{sec:results}
\blue{First, we demonstrate the forecasting performance of BiTCN on a set of real-world datasets to substantiate our claim that BiTCN achieves state-of-the-art forecasting performance with less parameters than competing methods (Section~\ref{subsec:fperformance}). 
Second, we will show the benefit of our architecture by studying the forecasting efficiency in terms of model complexity, training time and energy cost (Section~\ref{subsec:fefficiency}).
Finally, we study the impact of our design choices, such as employing the Student's t(3)-distribution (Section~\ref{subsec:studentteffect}), the effect of our forward-looking module (Section~\ref{subsec:fmodule}) and how BiTCN's performance is impacted by its hyperparameters (Section~\ref{subsec:hyperparams}).}

\subsection{Forecasting Effectiveness} \label{subsec:fperformance}
\blue{We report the forecasting performance in Table~\ref{tab:results}. Although similar model dimensions are used for each method across the experiments, parameter counts for each model may be different per experiment due to the size of the embedding dimension required to embed the categorical input vector \(\vec{a}_{cat}\), which has a different dimension for each dataset.}

\blue{We observe the following per dataset:
\begin{itemize}
	\item \texttt{Electricity}: BiTCN performs on par with the best methods WaveNet and TransformerConv.
	\item \texttt{Traffic}: BiTCN performs on par with the best methods.
	\item \texttt{Favorita}: BiTCN outperforms or performs similar to competing methods on all metrics.
	\item \texttt{WebTraffic}: BiTCN performs similar to best-performing method TCN on mean quantile loss and in line with other methods on the other metrics.
\end{itemize}
On average, BiTCN ranks highest at an overall average ranking of \(1.75\). Given these observations, we conclude that BiTCN can achieve state-of-the-art results on a set of real world probabilistic forecasting tasks whilst using significantly less parameters than the second best performing method, TransformerConv (average ranking of \(2.25\)). }

\blue{Secondly, even though not a primary objective of our paper, we confirm the findings from \cite{salinas_deepar_2019, li_enhancing_2019} that traditional methods such as Seasonal Naive, ETS and Theta are outperformed by neural network methods on the task of multistep probabilistic forecasting.}

\blue{Finally, we find that LightGBM performs reasonable but not as well as expected. We attribute this to our the experimental setup. First, to facilitate a like-for-like comparison to neural network-based methods, we use the same features for LightGBM as we provide to our neural networks. In practice, practitioners often spend a lot of time to engineer features that provide a better signal to a LightGBM model, which typically improves performance. Second, we apply the LightGBM model recursively, which means that any bias and variance that enters the model may be propagated to future timesteps. This issue is confirmed from results from the M5 competition, where most participants created separate LightGBM models for every timestep to avoid this bias/variance propagation. However, for our setting this would imply creating not only a separate model per quantile, but also per timestep. This would give LightGBM a somewhat unfair advantage compared to neural network based methods.}

\begin{table}
\caption{\blue{Forecasting results on various point and range accuracy metrics. For the neural network methods, we report mean metrics over 5 different seeds of parameter initializations per method, with standard deviation of the metric in brackets. Lower is better, bold indicates best method for the metric. Rank denotes the rank of the methods across the four metrics for each dataset (lower is better).}}
\label{tab:results}
\begin{center}
\blue{\begin{tabular}{l c  cccc c}
\toprule 
 &No.    & \multicolumn{2}{ c }{Point metrics}   & \multicolumn{2}{ c }{Range metrics} & Rank \\
 \cmidrule(r){3-4}\cmidrule(r){5-6}
Dataset/Method & parameters  & sMAPE & NRMSE & Q(0.5) & mQ & \\
\midrule
\texttt{Electricity} \\
\hspace{0.1cm} 	Seasonal Naive &0	&0.112	&0.719	&0.078	&0.078	&8 \\
\hspace{0.1cm} 	ETS &100k	&0.201	&1.755	&0.152	&0.136	&11 \\
\hspace{0.1cm} 	Theta &20k	&0.177	&1.575	&0.141	&0.164	&10 \\
\hspace{0.1cm} 	LightGBM &~2M	&0.089	&0.679	&0.065	&0.077	&5 \\
\hspace{0.1cm} 	DeepAR &45k	&0.099 (0.0044)	&0.671 (0.0142)	&0.069 (0.0010)	&0.057 (0.0009)	&7 \\
\hspace{0.1cm} 	ML-RNN &2.9M	&0.115 (0.0029)	&0.829 (0.0555)	&0.086 (0.0030)	&0.068 (0.0024)	&9 \\
\hspace{0.1cm} 	TCN &46k	&0.103 (0.0017)	&0.668 (0.0121)	&0.068 (0.0011)	&0.056 (0.0010)	&6 \\
\hspace{0.1cm} 	WaveNet &48k	&0.093 (0.0026)	&\textbf{0.646 (0.0095)}	&0.063 (0.0005)	&0.052 (0.0005)	&2 \\
\hspace{0.1cm} 	TransformerConv &415k	&\textbf{0.083 (0.0014)}	&0.658 (0.0237)	&\textbf{0.062 (0.0012)}	&0.052 (0.0010)	&1 \\
\hspace{0.1cm} 	\textit{BiTCN} &49k	&0.089 (0.0009)	&0.648 (0.0090)	&0.063 (0.0003)	&\textbf{0.052 (0.0003)}	&2 \\
\midrule
\texttt{Traffic} \\
\hspace{0.1cm} 	Seasonal Naive &0	&0.351	&0.667	&0.285	&0.285	&9 \\
\hspace{0.1cm} 	ETS &100k	&0.483	&0.693	&0.371	&0.348	&10 \\
\hspace{0.1cm} 	Theta &20k	&0.512	&3.413	&0.980	&0.970	&11 \\
\hspace{0.1cm} 	LightGBM &~2M	&0.128	&0.358	&0.111	&0.114	&4 \\
\hspace{0.1cm} 	DeepAR &57k	&0.179 (0.0182)	&0.408 (0.0274)	&0.131 (0.0141)	&0.111 (0.0126)	&8 \\
\hspace{0.1cm} 	ML-RNN &2.4M	&0.140 (0.0021)	&0.387 (0.0026)	&0.123 (0.0017)	&0.102 (0.0013)	&7 \\
\hspace{0.1cm} 	TCN &57k	&0.150 (0.0128)	&0.373 (0.0074)	&0.117 (0.0027)	&0.098 (0.0023)	&6 \\
\hspace{0.1cm} 	WaveNet &60k	&0.165 (0.0054)	&0.357 (0.0028)	&0.108 (0.0018)	&0.091 (0.0015)	&3 \\
\hspace{0.1cm} 	TransformerConv &372k	&0.130 (0.0035)	&\textbf{0.353 (0.0020)}	&\textbf{0.105 (0.0020)}	&\textbf{0.089 (0.0014)}	&1 \\
\hspace{0.1cm} 	\textit{BiTCN} &61k	&\textbf{0.127 (0.0005)}	&0.372 (0.0142)	&0.108 (0.0005)	&0.091 (0.0004)	&2 \\
\midrule
\texttt{Favorita} \\
\hspace{0.1cm} 	Seasonal Naive &0	&1.001	&4.541	&1.053	&1.053	&10 \\
\hspace{0.1cm} 	ETS &100k	&1.024	&4.297	&1.021	&0.842	&8 \\
\hspace{0.1cm} 	Theta &20k	&1.060	&2.369	&0.844	&0.922	&8 \\
\hspace{0.1cm} 	LightGBM &~2M	&0.879	&1.562	&0.489	&0.396	&5 \\
\hspace{0.1cm} 	DeepAR &61k	&\textbf{0.574 (0.1823)}	&1.680 (0.1040)	&0.543 (0.0216)	&0.422 (0.0150)	&5 \\
\hspace{0.1cm} 	ML-RNN &1.9M	&0.689 (0.0132)	&1.406 (0.0483)	&0.461 (0.0079)	&0.362 (0.0064)	&4 \\
\hspace{0.1cm} 	TCN &61k	&0.612 (0.0989)	&1.323 (0.0562)	&0.440 (0.0210)	&0.350 (0.0134)	&3 \\
\hspace{0.1cm} 	WaveNet &66k	&0.711 (0.0261)	&1.623 (0.1773)	&0.512 (0.0427)	&0.405 (0.0319)	&7 \\
\hspace{0.1cm} 	TransformerConv &210k	&0.673 (0.0046)	&1.319 (0.0602)	&0.439 (0.0179)	&\textbf{0.346 (0.0129)}	&1 \\
\hspace{0.1cm} 	\textit{BiTCN} &66k	&0.674 (0.0015)	&\textbf{1.317 (0.0179)}	&\textbf{0.432 (0.0049)}	&0.347 (0.0033)	&1 \\
\midrule
\texttt{WebTraffic} \\
\hspace{0.1cm} 	Seasonal Naive &0	&0.357	&4.709	&0.414	&0.414	&10 \\
\hspace{0.1cm} 	ETS &100k	&0.311	&4.096	&0.350	&0.359	&8 \\
\hspace{0.1cm} 	Theta &20k	&0.338	&4.132	&0.340	&0.476	&9 \\
\hspace{0.1cm} 	LightGBM &~2M	&0.260	&4.022	&0.273	&1.002	&6 \\
\hspace{0.1cm} 	DeepAR &238k	&0.279 (0.0054)	&4.003 (0.2526)	&0.282 (0.0060)	&0.244 (0.0050)	&5 \\
\hspace{0.1cm} 	ML-RNN &2.4M	&0.246 (0.0064)	&4.027 (0.0269)	&0.270 (0.0048)	&0.234 (0.0051)	&4 \\
\hspace{0.1cm} 	TCN &239k	&\textbf{0.234 (0.0034)}	&\textbf{3.718 (0.1692)}	&\textbf{0.253 (0.0048)}	&\textbf{0.231 (0.0046)}	&1 \\
\hspace{0.1cm} 	WaveNet &241k	&0.236 (0.0037)	&3.953 (0.3083)	&0.268 (0.0129)	&0.244 (0.0107)	&2 \\
\hspace{0.1cm} 	TransformerConv &630k	&0.246 (0.0065)	&4.191 (0.3368)	&0.301 (0.0225)	&0.273 (0.0198)	&6 \\
\hspace{0.1cm} 	\textit{BiTCN} &242k	&0.254 (0.0062)	&3.988 (0.2177)	&0.267 (0.0057)	&0.232 (0.0041)	&2 \\
\bottomrule
\end{tabular}}
\end{center}
\end{table}

\subsection{Forecasting Efficiency} \label{subsec:fefficiency}

\paragraph{Model complexity} As can be seen from Table~\ref{tab:results}, BiTCN requires almost an order of magnitude less parameters than the TransformerConv. 
This is an indication that our architecture is more efficient, and possibly a better choice for the task of probabilistic forecasting compared to existing neural architectures. To understand computational complexity, we are mainly interested in the sizes of the following parameters \blue{(we study the impact of varying these parameters in Section~\ref{subsec:fefficiency})}: 
\begin{itemize} 
\item \(N\), the number of layers of a neural network; 
\item \(k\), the kernel size of a convolution; 
\item \(T\), the sequence length used in the network; 
\item \(d_{h}\), the hidden dimension of the network; and 
\item \(d_{out}\), the number of output channels of a convolutional layer. 
\end{itemize}

The computational complexity of each of our TCN layers can be computed by adding the computational complexity of respectively the convolution layer, the activation, and the dense output layer:
\begin{align}
	O(k \cdot{T}\cdot{d_{h}}\cdot{d_{out}} + T^2 &+ d_{out} \cdot{d_{h}}\cdot{T^2} ) = \nonumber \\
	 		&O(k \cdot{T}\cdot{d_{h}^2} + d_{h}\cdot{T^2}), 
\end{align}
where in BiTCN, \(d_{out} = 4\cdot{d_{h}}\). Hence, the complexity for an \(N\)-layer network of our architecture is \( O( N\cdot (k \cdot{T}\cdot{d_{h}^2} + d_{h}\cdot{T^2})) \) with \(O(1)\) sequential operations per layer. In comparison, an LSTM-based architecture such as DeepAR has a computational complexity of \(O(N \cdot (T \cdot{d_{h}}^2))\), but requires \(O(T)\) sequential operations, which significantly slows down training when sequence length increases. 

Finally, Transformer-based architectures have a computational complexity of \(O(N \cdot (T \cdot{d_{h}}^2 + d_h\cdot{T}^2))\) and require \(O(1)\) sequential operations \cite{vaswani_attention_2017}, but through the use of dilated convolutions in the self-attention mechanism the computational complexity of the Transformer architecture of \cite{li_enhancing_2019} becomes \(O(N \cdot (k \cdot{T} \cdot{d_{h}}^2 + d_h\cdot{T}^2))\), with \(k\) the kernel size of the dilated convolutions. Hence, the computational complexity of the TransformerConv architecture is equal to that of BiTCN, however in practice this leads to different outcomes. Even though BiTCN requires more layers to ensure a sufficient receptive field, it requires a much smaller hidden dimension throughout the network. In contrast, the TransformerConv network requires less layers but a higher hidden dimension in order to support sufficient model capacity for each of the attention heads. 

\paragraph{Model complexity compared to non-neural methods} \blue{We find that BiTCN requires signficantly less parameters than LightGBM. What causes this? Suppose we would like to have a probabilistic forecast for 28 days, for 9 quantiles. This requires at least 9 models in the GBT setting. In our setting, each GBT consists of approximately 2000 trees, and is trained with a maximum number of leaf nodes of 127. Hence, this yields 254k parameters. Then, we have 9 such models, yielding approximately 2M parameters. If one would opt for separate models for each forecast day, one would require $28\cdot2M = 64M$ parameters, which is orders of magnitude larger than comparable NN-based solutions. What causes this difference in model size? Ultimately, this is due to the fact that NN methods are better representational learning methods, or stated otherwise, these methods better learn to compress the data. In a NN-based solution, an existing set of parameters is modified during learning. GBTs however, work incrementally, and add parameters for each iteration. Hence, these models do not compress the data as much as the NN does, yielding models that have a higher complexity. }

\begin{figure}
\centering
\centerline{\includegraphics[width=\textwidth]{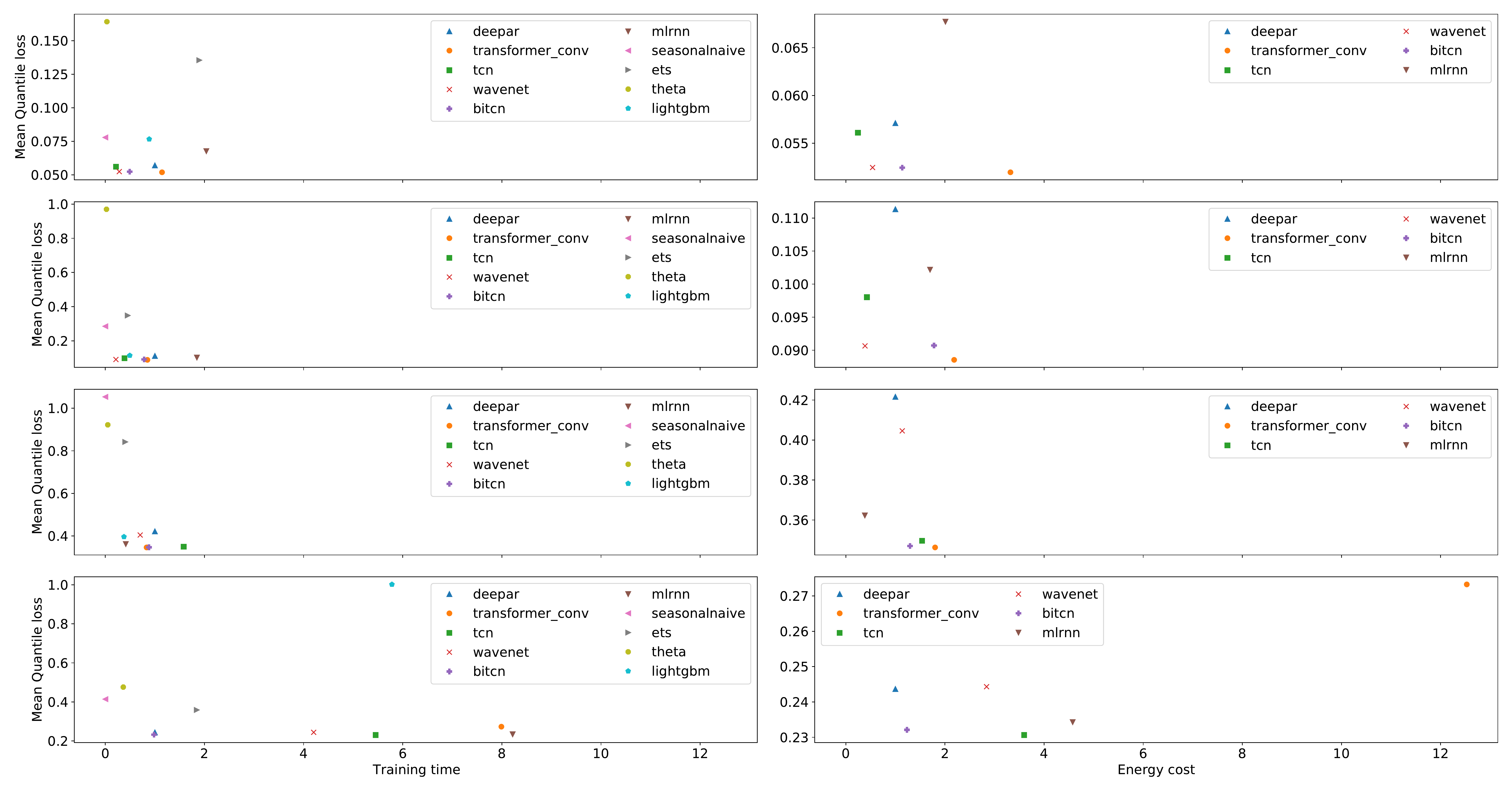}}
\caption{\blue{Running time (left) and energy cost (right) compared to mean quantile loss on the test set for respectively \texttt{Electricity} (top), \texttt{Traffic}, \texttt{Favorita} and \texttt{Webtraffic} (bottom) datasets, where DeepAR is the baseline.}}
\label{fig:cost}
\end{figure}

\paragraph{Training time} \blue{One of the benefits of a model with less parameters is that \textit{ceteris paribus} it can be trained faster. Therefore, we compare training times for each method in the left plane of Figure~\ref{fig:cost}. For each method, the left side of Figure~\ref{fig:cost} shows the mean quantile loss on the test set as compared against the training time. The training times are normalized against a DeepAR baseline. We observe that BiTCN is the only algorithm that sits in the lower left corner - which implies smallest training time with the lowest quantile loss on the test set - for each of the datasets. Also, we observe that LightGBM's training time is comparable to those of NN methods, which is mostly due to its requirement to create separate models for each individual quantile in the forecast. Note that the latter finding is subject to hardware considerations, as the LightGBM models are trained on CPU whilst the neural models are trained on a GPU.}

\paragraph{Energy cost} \blue{Even though one method may train faster than the other, the energy cost for training may still exceed the energy cost of training other models, due to an increased resource consumption. Therefore, we analyze the consumed energy for training each model in the right half plane of Figure~\ref{fig:cost}. For this experiment, we ran each method separately and sequentially for a single epoch, and measured the average GPU board power draw of a nVidia GTX 1080Ti using GPU-Z. During each experiment, we fixed the amount of CPU threads available to the training process and there were no other processes running on the GPU used for measuring energy cost. The obtained result indicates the total average amount of energy in Joules that is required to compute a single epoch for each method. This energy consumption is then multiplied with the average number of epochs (across the 5 seeds) required to train each method to obtain the overall GPU energy cost. Finally, we normalize this energy cost against a DeepAR baseline. BiTCN is the strongest performer across all the datasets, as it sits in the lower left corner on each graph. We now see a more clear distinction between TransformerConv and the other methods, and its performance is now less favourable, as it requires a higher energy consumption on every dataset compared to the other methods, and we see energy consumption differences of 20\% up to an order of magnitude compared to BiTCN. In practice, this result is beneficical for practitioners who are interested in optimizing electricity and cooling costs, which commonly represent a significant portion of data center operating cost \cite{strubell_energy_2020}.}

\paragraph{Memory usage} The Transformer's memory consumption scales quadratically with sequence length \cite{li_enhancing_2019}, as all activations of the multi-head attention layers need to be stored during a forward pass of the network to use during the backward pass. In contrast, for temporal convolutional architectures memory consumption predominantly scales with 
\begin{enumerate*}[label=(\arabic*)]
\item the number of layers required to obtain a sufficient receptive field, and 
\item the number of parameters of these layers. 
\end{enumerate*}
This enables training these models using less memory, which in turn enables the use of cheaper resources. In Table~\ref{tab:memoryconsumption} we show the GPU memory consumption during training compared between BiTCN and TransformerConv across datasets. When comparing similar batch sizes, we observe a difference in memory consumption of 2--4x. In practice, this means that BiTCN models can be trained on much cheaper GPUs. For example, a TransformerConv model with batch size 512 requires a high-end video card such as the nVidia RTX 2080Ti (which is equipped with 11GB of GPU memory) to train the model on the \texttt{Electricity} and \texttt{Traffic} datasets, whereas a similar batch size BiTCN model would only require a low-end RTX 2060. The first card typically retails for over USD 1000, whereas the latter can be acquired starting from USD 300 on Amazon. 

\begin{table}[t]
\caption{GPU memory consumption in Gigabytes during training of BiTCN compared to TransformerConv.}
\label{tab:memoryconsumption}
\begin{center}
\begin{tabular}{l cccc}
\toprule 
Batch size / Dataset & BiTCN & TransformerConv & Ratio (x) \\
\midrule
\texttt{128} \\
\hspace{0.1cm} \texttt{Electricity} &1.54 & 3.43 &2.23 \\
\hspace{0.1cm} \texttt{Traffic} &1.52 & 3.37 &2.21 \\
\hspace{0.1cm} \texttt{Favorita} &1.34 & 2.15 &1.60 \\
\hspace{0.1cm} \texttt{WebTraffic} &1.36 & 2.20 &1.61 \\
\texttt{256} \\
\hspace{0.1cm} \texttt{Electricity} &1.91 & 6.61 &3.46 \\
\hspace{0.1cm} \texttt{Traffic} &1.90 & 6.55 &3.45 \\
\hspace{0.1cm} \texttt{Favorita} &1.49 & 3.47 &2.33 \\
\hspace{0.1cm} \texttt{WebTraffic} &1.51 & 3.46 &2.30 \\
\texttt{512} \\
\hspace{0.1cm} \texttt{Electricity} &2.62 & 10.20 &3.89 \\
\hspace{0.1cm} \texttt{Traffic} &2.60 & 10.18 &3.91 \\
\hspace{0.1cm} \texttt{Favorita} &1.77 & 5.95 &3.36 \\
\hspace{0.1cm} \texttt{WebTraffic} &1.78 & 5.95 &3.33 \\
\bottomrule
\end{tabular}
\end{center}
\end{table}

\subsection{Effect of Student's t(3)-distribution} \label{subsec:studentteffect}
\begin{figure}[t]
\centering
\centerline{\includegraphics[width=0.5\columnwidth]{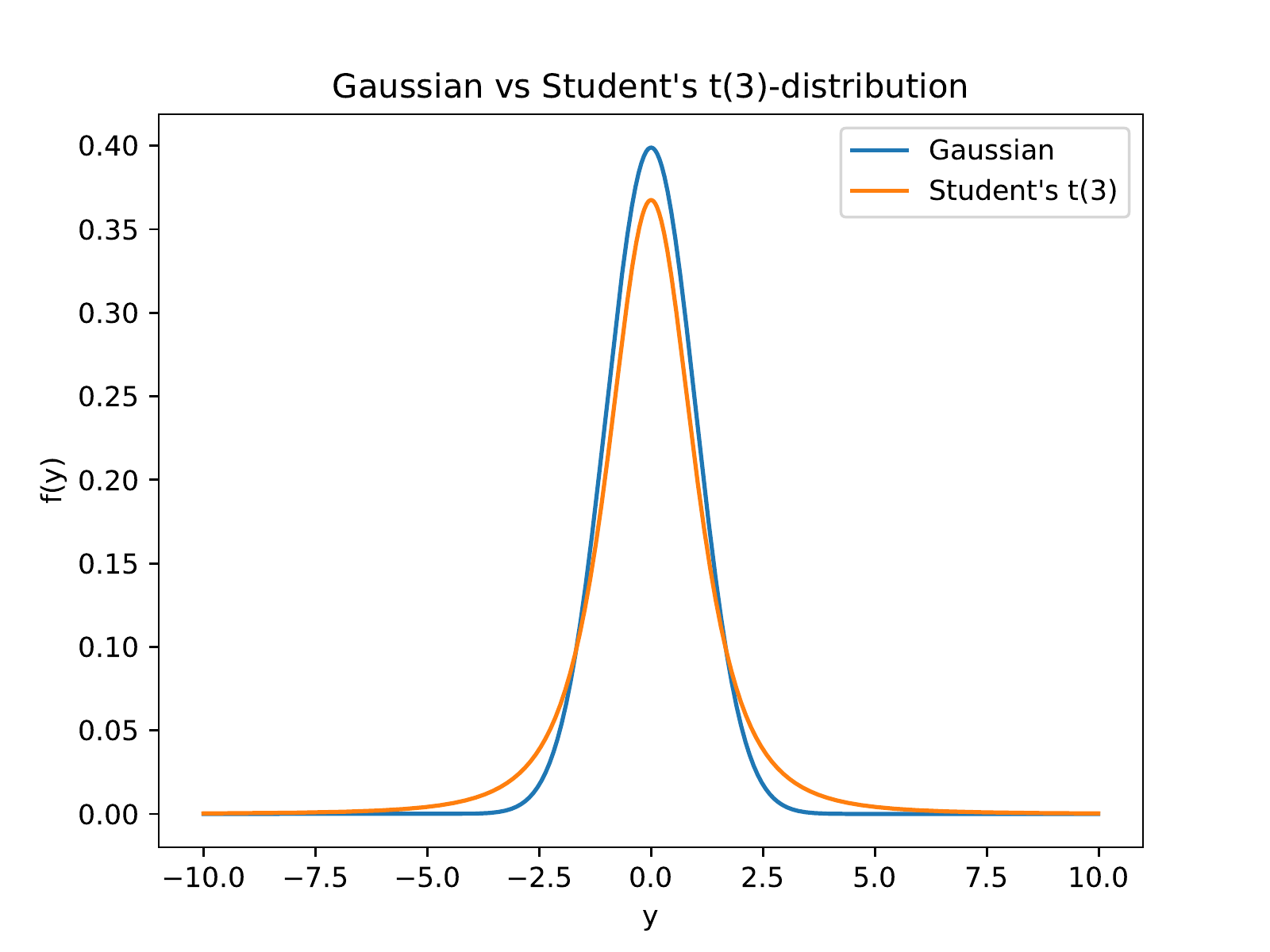}}
\caption{The probability density function of the normal distribution (Gaussian with (loc, scale) = (0,1)) compared to the Student's t(3)-distribution with (loc, scale) = (0,1). The normal distribution has a thin tail compared to the Student's t(3)-distribution.}
\label{fig:distributions}
\end{figure}

\blue{To illustrate the benefits of using a Student's t(3)-distribution for probabilistic forecasting, we re-run the experiments from Section~\ref{subsec:fperformance} on the \texttt{Electricity} and \texttt{Traffic} datasets for BiTCN using a parameterized Gaussian output distribution.  
We keep the same training settings as before, however we note that the Gaussian distribution in our forecasting setting requires clipping gradients to achieve a stable training regime. This requires tuning yet another hyperparameter -- the maximum gradient norm. Why is this clipping necessary? It is a direct consequence of the `thin-tailedness' of the probability density function of the Gaussian, which is illustrated in Figure~\ref{fig:distributions}. The thin tail of the Gaussian causes the log-probability during training to exert numerical instability. On the contrary, a fat-tailed distribution such as the Student's t(3)-distribution enables a stabler training regime. Aside from this practical disadvantage, a thin-tailed distribution is also expected to perform worse on forecasting for processes that do not follow a normal distribution in their output data. We confirm this expectation by observing the results in Table~\ref{tab:gaussian}, where we compare the forecasting performance with a Gaussian as output distribution vis-a-vis the Student's t(3)-distribution. On both \texttt{Electricity} and \texttt{Traffic} datasets, we see performance differences of 5-15\% when using a Gaussian output distribution instead of a Student's t(3)-distribution. For the the \texttt{Traffic}, the differences are generally larger, which is expected as this dataset is relatively skewed towards zero and hence benefits more from using an output distribution that has a heavier tail such as the Student's t(3)-distribution. 
Finally, we also observe lower variance in our test scores for the Student's t(3) distribution, which indicates a more stable training regime for this loss function. Our conclusion from this experiment is that the Student's t(3)-distribution improves forecasting performance whilst removing a hyperparameter from the optimization problem and enabling a more stable training regime. }

\begin{table}
\caption{\blue{Forecasting results on various point and range accuracy metrics comparing Student's t(3)-distribution with the Gaussian distribution as loss function. Lower is better, bold indicates best loss function for the method. We report mean metrics over 5 different seeds per method with standard deviation in brackets.}}
\label{tab:gaussian}
\begin{center}
\blue{\begin{tabular}{l cc  cc}
\toprule 
 & \multicolumn{2}{ c }{Point metrics}   & \multicolumn{2}{ c }{Range metrics} \\
 \cmidrule(r){2-3}\cmidrule(r){4-5}
Dataset / Method &  sMAPE & NRMSE & Q(0.5) & mQ \\
\midrule
\texttt{Electricity} \\
\hspace{0.1cm} 	BiTCN \textit{Gaussian}		&0.117 (0.0054)  &0.713 (0.0280)  &0.076 (0.0036) &0.062 (0.0023) \\
\hspace{0.1cm}  BiTCN \textit{Student's t(3)} 	&\textbf{0.089 (0.0009)}	&\textbf{0.648 (0.0090)}	&\textbf{0.063 (0.0003)}	&\textbf{0.052 (0.0003)}\\
\midrule
\texttt{Traffic} \\
\hspace{0.1cm} 	BiTCN \textit{Gaussian}		&0.175 (0.0089) &0.404 (0.0763) & 0.141 (0.0016) &0.115 (0.0012) \\
\hspace{0.1cm} 	BiTCN \textit{Student's t(3)}	&\textbf{0.127 (0.0005)}	&\textbf{0.372 (0.0142)}	&\textbf{0.108 (0.0005)}	&\textbf{0.091 (0.0004)} \\
\bottomrule
\end{tabular}}
\end{center}
\end{table}

\subsection{\blue{Effect of forward-looking module}} \label{subsec:fmodule}
\blue{To study the effect of our forward-looking module, we re-run the experiments from Section~\ref{subsec:fperformance} on every dataset where we disable the forward-looking module. We report the results in Table~\ref{tab:forwardmodule}. We observe a positive impact of about 0-2\% from the forward-module in all but one dataset, and especially in the \texttt{Traffic} and \texttt{Favorita} datasets. Especially for the \texttt{Favorita} dataset this is expected, as this dataset provides very informative future covariates (e.g. whether there is a holiday on a particular day). On the contrary, the other datasets mostly contain covariates related to the day of the week or day of the month, which are less informative. }

\begin{table}
\caption{\blue{Forecasting results on point and range accuracy metrics, ablating for our forward-looking module. We report mean metrics over 5 different seeds of parameter initializations per method, with standard deviation of the metric in brackets. Lower is better, bold indicates best method for the metric.}}
\label{tab:forwardmodule}
\begin{center}
\blue{\begin{tabular}{l c  cccc}
\toprule 
 &No.    & \multicolumn{2}{ c }{Point metrics}   & \multicolumn{2}{ c }{Range metrics}  \\
 \cmidrule(r){3-4}\cmidrule(r){5-6}
Dataset/Method & parameters  & sMAPE & NRMSE & Q(0.5) & mQ \\
\midrule
\texttt{Electricity} \\
\hspace{0.1cm} 	BiTCN &49k	&\textbf{0.089 (0.0009)}	&0.648 (0.0090)	&\textbf{0.063 (0.0003)}	&\textbf{0.052 (0.0003)} \\
\hspace{0.1cm} 	BiTCN (w/o forward) &40k	&0.089 (0.0012)	&\textbf{0.647 (0.0050)}	&0.064 (0.0005)	&0.053 (0.0004) \\
\midrule
\texttt{Traffic} \\
\hspace{0.1cm} 	BiTCN &61k	&\textbf{0.127 (0.0005)}	&\textbf{0.372 (0.0142)}	&\textbf{0.108 (0.0005)}	&\textbf{0.091 (0.0004)} \\
\hspace{0.1cm} 	BiTCN (w/o forward) &52k	&0.128 (0.0007)	&0.546 (0.3427)	&0.112 (0.0062)	&0.094 (0.0047) \\
\midrule
\texttt{Favorita} \\
\hspace{0.1cm} 	BiTCN &66k	&\textbf{0.674 (0.0015)}	&\textbf{1.317 (0.0179)}	&\textbf{0.432 (0.0049)}	&\textbf{0.347 (0.0033)} \\
\hspace{0.1cm} 	BiTCN (w/o forward) &58k	&0.683 (0.0024)	&1.390 (0.0306)	&0.460 (0.0084)	&0.366 (0.0058) \\
\midrule
\texttt{WebTraffic} \\
\hspace{0.1cm} 	BiTCN &242k					&0.254 (0.0062)	&\textbf{3.988 (0.2177)}	&0.267 (0.0057)	&0.232 (0.0041) \\
\hspace{0.1cm} 	BiTCN (w/o forward) &233k	&\textbf{0.252 (0.0055)}	&4.004 (0.1903)	&\textbf{0.265 (0.0046)}	&\textbf{0.232 (0.0032)} \\
\bottomrule
\end{tabular}}
\end{center}
\end{table}

\subsection{\blue{Effect of hyperparameters}} \label{subsec:hyperparams}
\blue{Finally, we briefly study the impact of the choice of key hyperparameters of BiTCN. We reran a set of experiments for several choices of hyperparameters on the \texttt{Electricity} and \texttt{Traffic} dataset for which we display the results in Table~\ref{tab:hypersens}. We highlight a number of interesting observations:
\begin{itemize}
	\item Probabilistic forecasting performance of BiTCN seems relatively robust against a wide set of hyperparameter choices, as we commonly observe differences of 0-5\% in mean quantile loss when varying hyperparameters, and BiTCN would rank as a top performer among the competing methods in Table~\ref{tab:results} for nearly all of the various hyperparameter settings.
	\item BiTCN's performance improves when the hidden dimension \(d_h\) (and thus the number of parameters) is increased. Conversely, performance also significantly degrades when the hidden dimension is reduced. However, an increased hidden size can result in both higher (\texttt{Electricity}) and lower (\texttt{Traffic}) training time and energy cost, which is due to the experiment requiring more (\texttt{Electricity}) and less (\texttt{Traffic}) epochs when increasing the hidden dimension. Also, the increased running time and energy cost is still less than that of the TransformerConv (ref. Figure~\ref{fig:cost}), further demonstrating that our architecture achieves the same performance but does so more efficiently.
	\item The dropout rate \(p_d\) seems very important, as excluding it by setting it to zero results in a large performance hit for both datasets.
	\item It seems beneficial to increase the kernel size \(k\) of the convolutions, as performance generally increases when \(k\) is higher.
\end{itemize}}

\begin{table}
\caption{\blue{Sensitivity of sMAPE, mean quantile loss, training time and energy cost of BiTCN when varying key hyperparameters: the batch size \(b_s\), the hidden size \(d_h\), the kernel size \(k\), the number of layers \(N\) and the dropout rate \(p_d\). For each row, only the value listed is changed with respect to the base case.}}
\label{tab:hypersens}
\begin{center}
\blue{\begin{tabular}{l  ccccccc ccc}
\toprule 
 &No.    &\(b_s\) &\(d_h\) &\(k\) & \(N\) &\(p_d\) &sMAPE & mQ &Training & Energy\\
 &parameters &&&&&&&&time&cost \\
\midrule
\texttt{Electricity} \\
\hspace{0.1cm} 	Base case &49k &512 &12 &9 &5 &0.1 &0.089	&0.052 &1.00 &1.00 \\
\hspace{0.1cm} 	 		 &  &256  & & & & &0.091 (2.4\%)	&0.054 (2.7\%)	&0.82	&0.75 \\
\hspace{0.1cm} 	 		 &  &128  & & & & &0.091 (2.1\%)	&0.053 (1.2\%)	&0.94	&0.69 \\
\hspace{0.1cm} 	 		 &166k  &  &24 & & & &0.083 (-6.0\%)	&0.051 (-1.9\%)	&1.54	&1.82 \\
\hspace{0.1cm} 	 		 &19k  &  &6 & & & &0.096 (8.1\%)	&0.054 (2.4\%)	&0.77	&0.68 \\
\hspace{0.1cm} 	 		 &39k  &  & &3 &7 & &0.095 (7.4\%)	&0.055 (4.6\%)	&0.48	&0.46 \\
\hspace{0.1cm} 	 		 &42k  &  & &5 &6 & &0.090 (1.6\%)	&0.053 (1.5\%)	&0.76	&0.73 \\
\hspace{0.1cm} 	 		 &50k  &  & &7 &6 & &0.089 (0.3\%)	&0.053 (0.9\%)	&0.78	&0.76 \\
\hspace{0.1cm} 	 		 &55k  &  & &11 &5 & &0.087 (-2.3\%)	&0.053 (0.2\%)	&1.00	&1.02 \\
\hspace{0.1cm} 	 		 & &   & & & &0.0 &0.101 (13.2\%)	&0.054 (2.1\%)	&0.62	&0.60 \\
\hspace{0.1cm} 	 		 & &   & & & &0.2 &0.093 (5.3\%)	&0.054 (2.2\%)	&0.71	&0.71 \\
\hspace{0.1cm} 	 		 & &   & & & &0.3 &0.094 (5.6\%)	&0.053 (1.8\%)	&1.14	&1.09 \\
\midrule
\texttt{Traffic} \\
\hspace{0.1cm} 	Base case &61k &512 &12 &9 &5 &0.1 &0.127	&0.091 & 1.00 & 1.00 \\
\hspace{0.1cm} 	 		 &  &256  & & & & &0.128 (0.6\%)	&0.090 (-1.0\%)	&1.14	&1.05 \\
\hspace{0.1cm} 	 		 &  &128  & & & & &0.126 (-0.6\%)	&0.090 (-1.2\%)	&1.56	&1.14 \\
\hspace{0.1cm} 	 		 &178k  &  &24 & & & &0.126 (-1.2\%)	&0.088 (-2.7\%)	&0.88	&1.02 \\
\hspace{0.1cm} 	 		 &31k  &  &6 & & & &0.138 (8.3\%)	&0.099 (9.2\%)	&0.95	&0.79 \\
\hspace{0.1cm} 	 		 &51k  &  & &3 &7 & &0.128 (1.0\%)	&0.092 (1.2\%)	&1.34	&1.25 \\
\hspace{0.1cm} 	 		 &54k  &  & &5 &6 & &0.127 (-0.2\%)	&0.09 (-0.8\%)	&1.34	&1.24 \\
\hspace{0.1cm} 	 		 &62k  &  & &7 &6 & &0.126 (-1.2\%)	&0.089 (-2.1\%)	&1.22	&1.23 \\
\hspace{0.1cm} 	 		 &67k  &  & &11 &5 & &0.126 (-0.5\%)	&0.093 (3.0\%)	&0.91	&0.95 \\
\hspace{0.1cm} 	 		 & &   & & & &0.0 &0.213 (67.7\%)	&0.143 (57.9\%)	&0.53	&0.55 \\
\hspace{0.1cm} 	 		 & &   & & & &0.2 &0.13 (1.9\%)	&0.093 (2.5\%)	&1.05	&1.06 \\
\hspace{0.1cm} 	 		 & &   & & & &0.3 &0.134 (5.8\%)	&0.096 (6.2\%)	&0.78	&0.77 \\
\bottomrule
\end{tabular}}
\end{center}
\end{table}

\section{Conclusion and Future Work}
In this work, we set out to find more parameter efficient methods of probabilistic forecasting. We hypothesized that by 
\begin{enumerate*}[label=(\arabic*)] 
\item smartly leveraging future covariate information often available in real-world settings, \item using a simple convolutional architecture, and 
\item employing a Student's t(3)-distribution,\end{enumerate*} 
it is possible to achieve state-of-the-art probabilistic forecasting performance compared to existing Transformer-based methods whilst requiring significantly less parameters. 
\blue{We find that our method, Bidirectional Temporal Convolutional Network (BiTCN), confirms these expectations, as we observe state-of-the-art forecasting effectiveness on a set of real-world benchmarks, even though BiTCN (i) uses an order of magnitude less parameters than the second-best Transformer-based method, (ii) requires at least 20\% less energy and (iii) about a quarter of the amount of memory to train the model on a GPU.}

We believe that these findings qualify BiTCN as a generic probabilistic forecasting method among practitioners, due to its simplicity and computational efficiency. 

Even though we observed the benefit of encoding future information to condition the current forecast on, the effect was relatively limited and even absent in some scenarios. 
For future work, we would therefore like to further investigate the benefit of this part of BiTCN by applying BiTCN in an industrial retail environment with thousands of products and stores, where a large set of historical data and future covariate information is available to condition a forecast on. An example of such setting is the M5 forecasting dataset \cite{makridakis_m5_2020}.
Secondly, we aim to investigate how our method scales to very long sequences (e.g., as in speech generation problems or high-frequency trading problems), where we expect to see more benefit of the forward-looking module. Finally, we intend to investigate creating a richer set of output distributions, in line with the recent work by \cite{gasthaus_probabilistic_2019}, which would further generalize our method by removing the choice of an output distribution.

\section*{Acknowledgments}
This research was (partially) funded by the Hybrid Intelligence Center, a 10-year program funded by the Dutch Ministry of Education, Culture and Science through the Netherlands Organisation for Scientific Research.\footnote{\url{https://www.hybrid-intelligence-centre.nl/}}

All content represents the opinion of the authors, which is not necessarily shared or endorsed by their respective employers and/or sponsors.

We thank the reviewers for their constructive feedback and help on improving our work.

\clearpage


\appendix

\section{Supplemental Materials}
\label{app:supplemental}
Table~\ref{tab:datasets} contains detailed descriptions of the datasets used in the paper.

\begin{sidewaystable*}
\caption{Dataset descriptions}
\label{tab:datasets}
\centering
\begin{tabular}{l c c c c c}
\toprule 
   & & \texttt{Electricity} & \texttt{Traffic} & \texttt{Favorita} &\texttt{WebTraffic} \\
\midrule
time series & \# & 370 & 963 & 170k & 10k \\
time series description & &customers &traffic lanes &item-store combinations &wikipedia pages \\
target & & \(\mathbb{R}^+\) &\([0,1]\) &\(\mathbb{R}^+\) &\(\mathbb{R}^+\) \\
train samples & \# & 500k & 500k & 500k & 500k \\
validation samples & \# & 7k & 7k & 10k &7k \\
test samples & \# & 7k & 7k & 10k &7k \\
time step & \(t\) & hour & hour & day & day  \\
input sequence length & \(t_0\) & 168 & 168 & 90 & 90 \\
output sequence length & \(T - t_0\) & 24 & 24 & 30 & 30 \\
covariate sequence length & \(T_c\) &500 &500 &150 &150 \\
categorical covariates & \# &1 &1 &2 &1 \\
embedding dimension & \(d_{emb}\) & 20 & 20 & [8, 3] &20 \\
numerical covariates & \(d_{cov}\) &7 &5 &7 &8 \\
lagged inputs & \(d_{lag}\) &1 &1 &1 &1 \\  
categorical covariate description & &\makecell{customer\_id} &\makecell{lane\_id} &\makecell[ct]{item\_id \\ store\_id} & page\_id \\
numerical covariates description & &\makecell[ct]{Month\_sin \\ Month\_cos \\ DayOfWeek\_sin \\ DayOfWeek\_cos \\ HourOfDay\_sin \\ HourOfDay\_cos \\ Online} &\makecell[ct]{DayOfWeek\_sin \\ DayOfWeek\_cos \\ HourOfDay\_sin \\ HourOfDay\_cos \\ Available} &\makecell[ct]{Holiday \\ On\_promotion \\ On\_sale \\ DayOfWeek\_sin \\ DayOfWeek\_cos \\ Month\_sin \\ Month\_cos} & \makecell[ct]{DayOfWeek\_sin \\ DayOfWeek\_cos \\ DayOfMonth\_sin \\ DayOfMonth\_cos \\ Month\_sin \\ Month\_cos \\ WeekOfYear\_sin \\ WeekOfYear\_cos}  \\
lagged input description & & target\_lagged &target\_lagged &target\_lagged & target\_lagged \\
\bottomrule
\end{tabular}
\end{sidewaystable*}

\begin{figure*}[ht]
\centering
\centerline{\includegraphics[width=\textwidth]{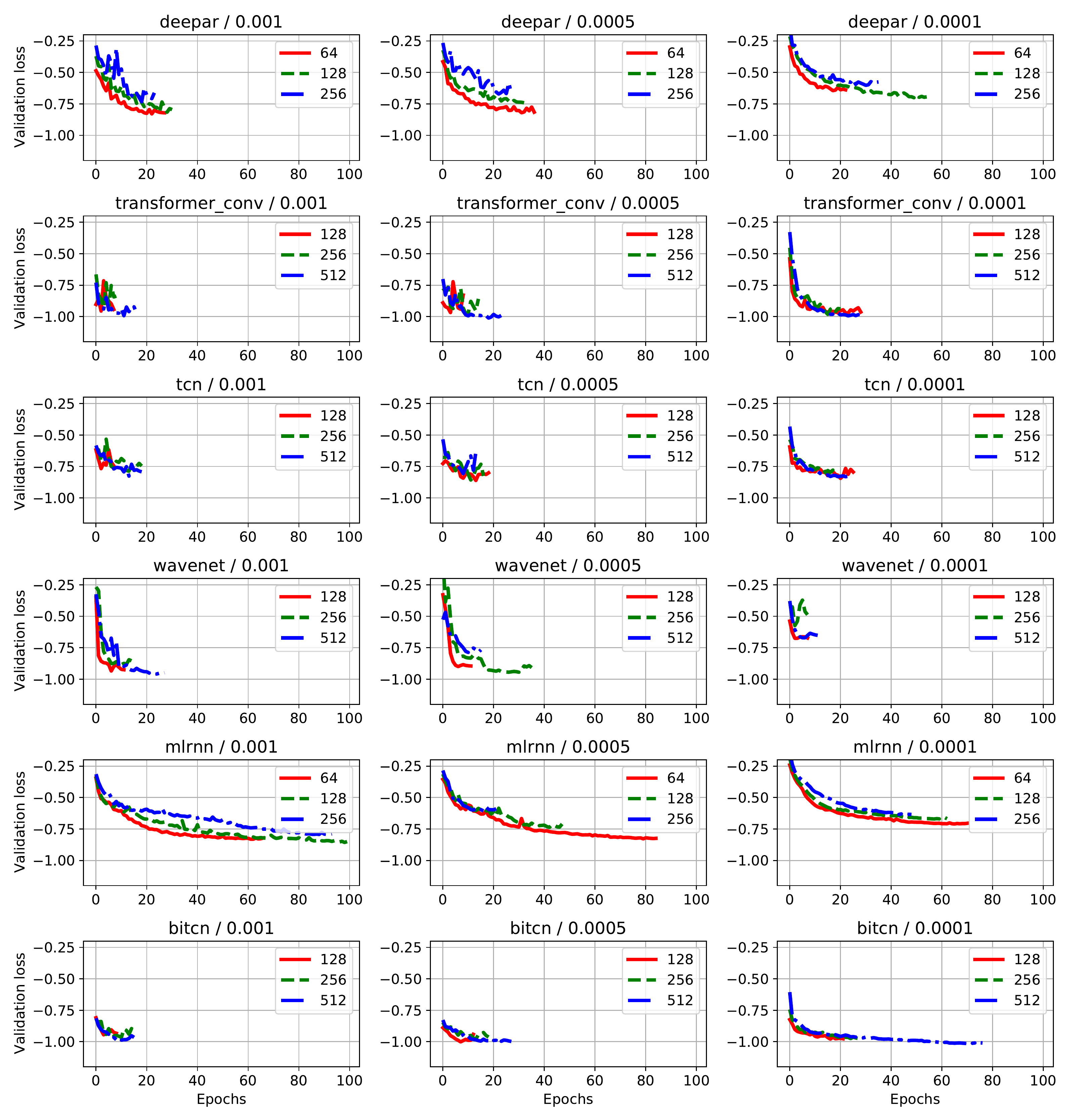}}
\caption{Training epochs versus validation loss for each method and learning rate for the \texttt{Electricity} dataset. The legends in the graphs denote the tested batch sizes. }
\label{fig:hyperparam_electricity}
\end{figure*}

\begin{figure*}[ht]
\centering
\centerline{\includegraphics[width=\textwidth]{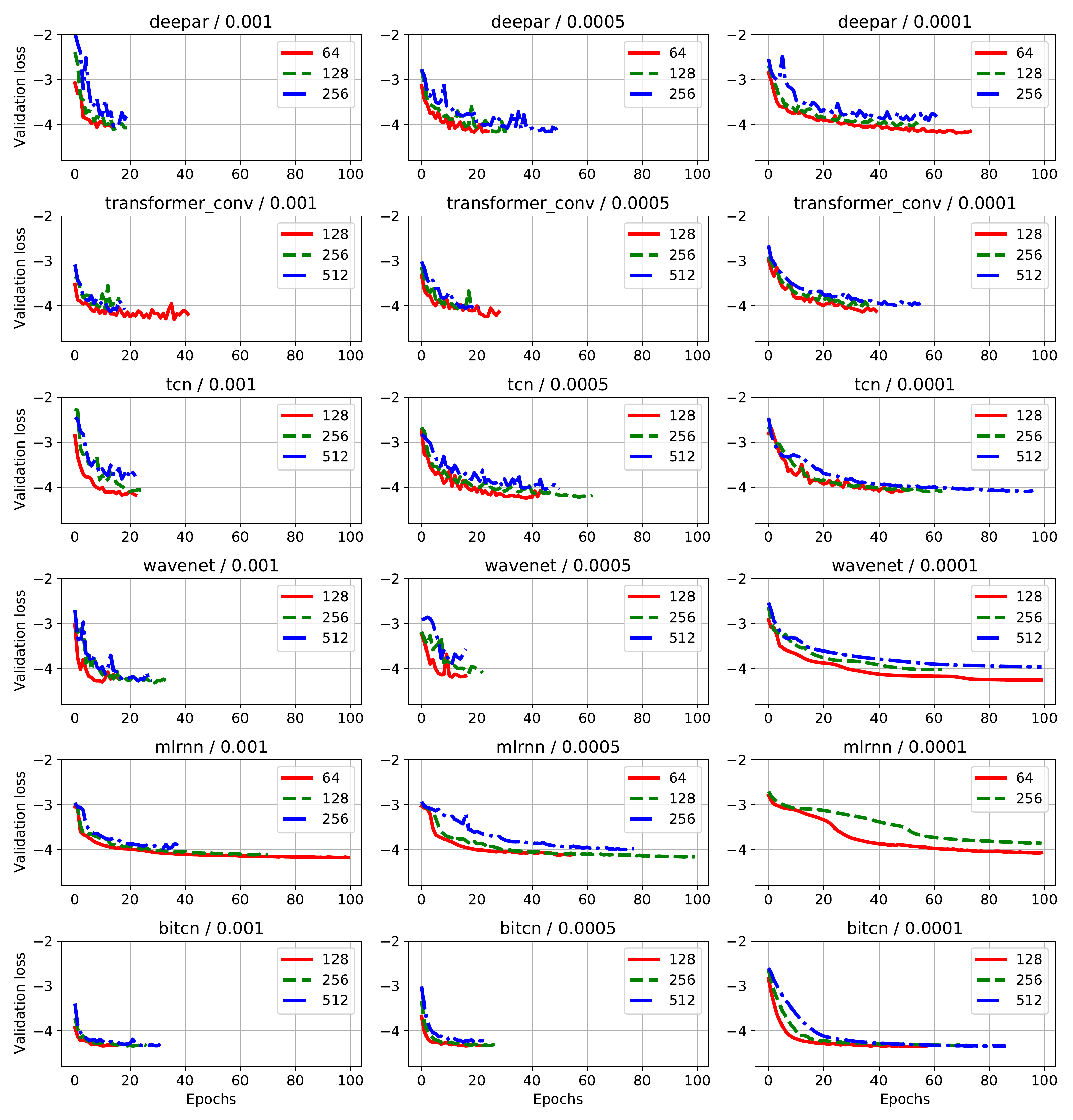}}
\caption{Training epochs versus validation loss for each method and learning rate for the \texttt{Traffic} dataset. The legends in the graphs denote the tested batch sizes. For ML-RNN, batch size 128 with learning rate 0.0001 resulted in repeated errors, therefore it is omitted in this graph.}
\label{fig:hyperparam_traffic}
\end{figure*}

\begin{figure*}[ht]
\centering
\centerline{\includegraphics[width=\textwidth]{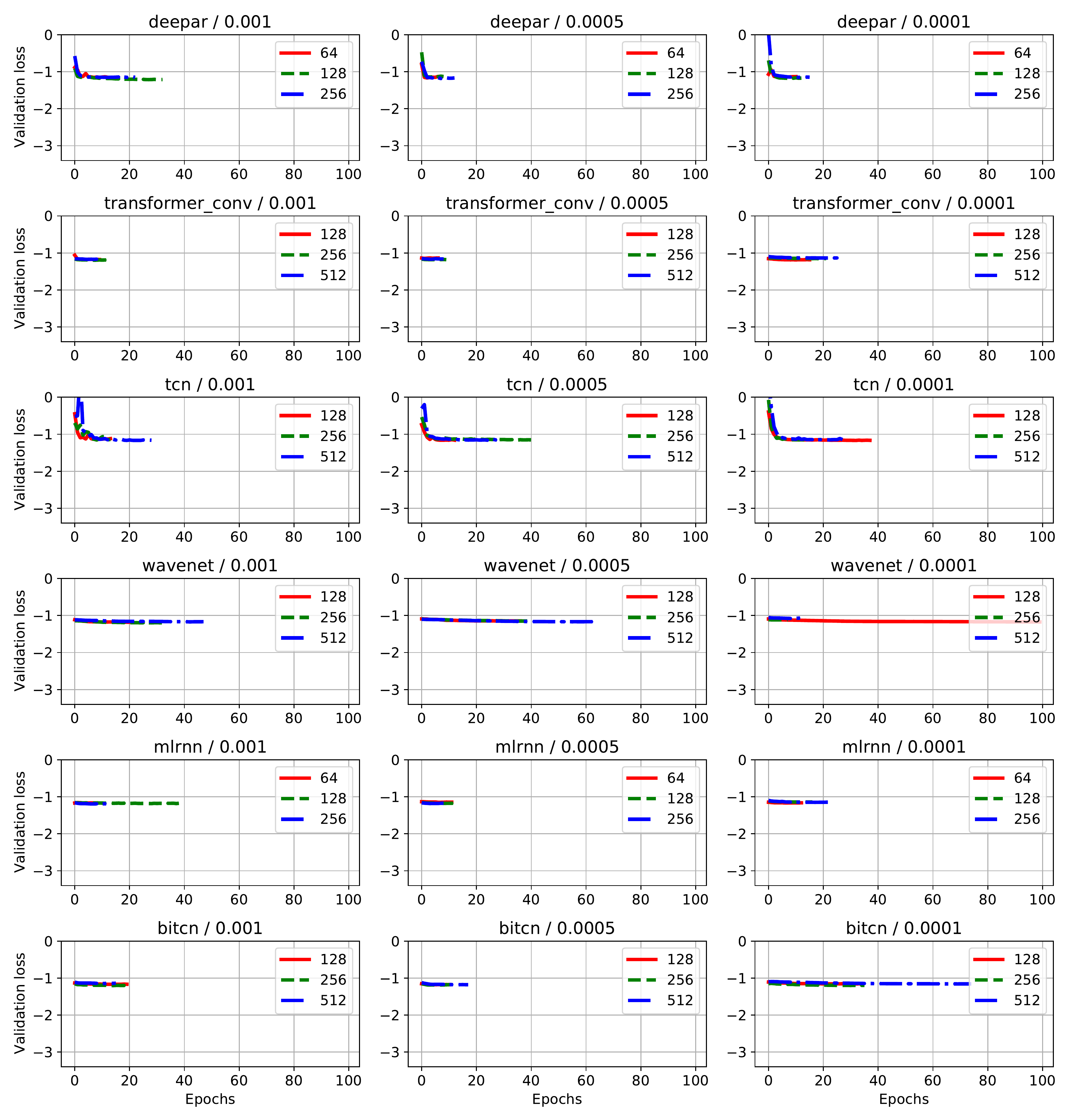}}
\caption{Training epochs versus validation loss for each method and learning rate for the \texttt{Favorita} dataset. The legends in the graphs denote the tested batch sizes.}
\label{fig:hyperparam_favorita}
\end{figure*}

\begin{figure*}[ht]
\centering
\centerline{\includegraphics[width=\textwidth]{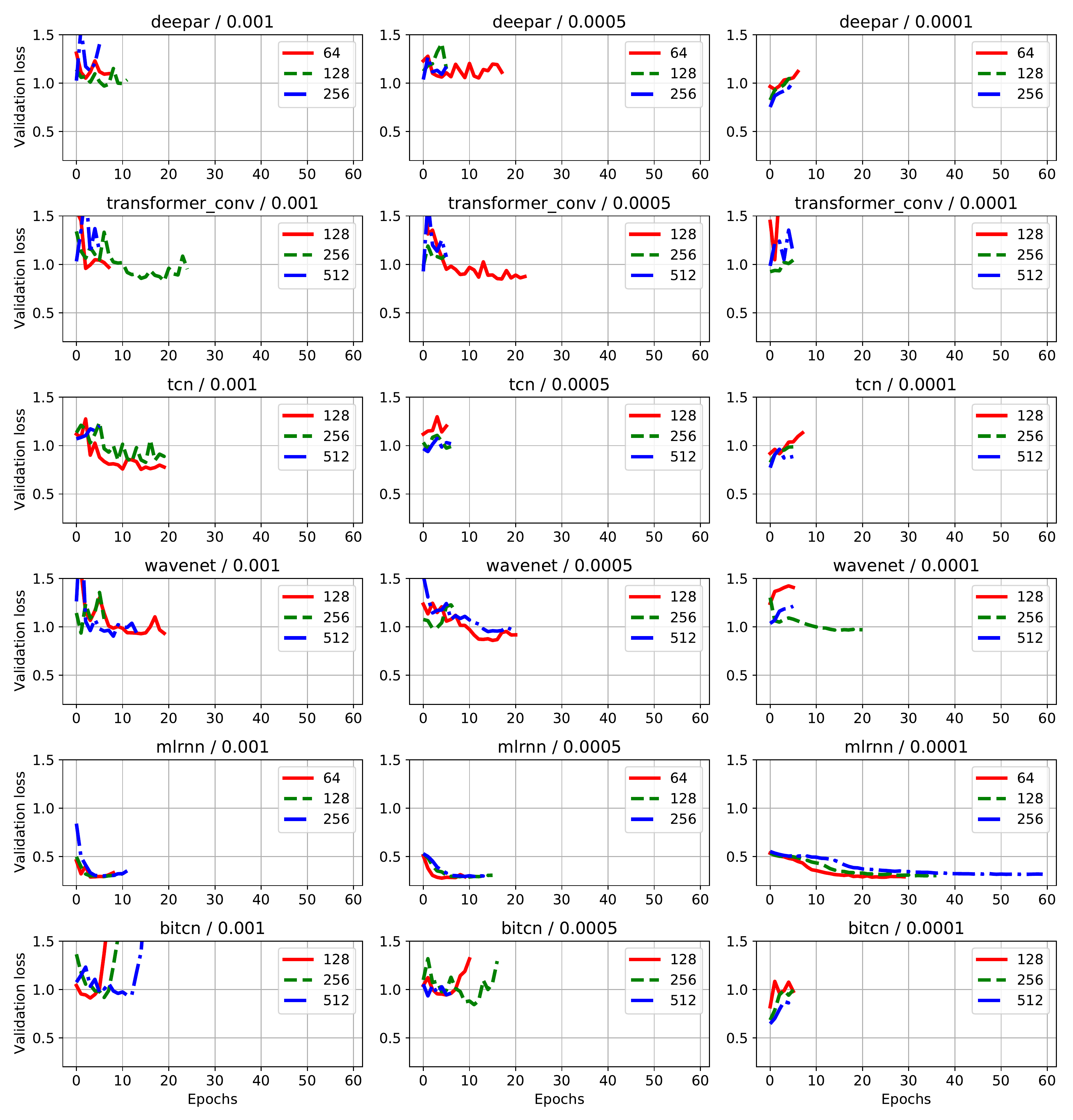}}
\caption{Training epochs versus validation loss for each method and learning rate for the \texttt{WebTraffic} dataset. The legends in the graphs denote the tested batch sizes.}
\label{fig:hyperparam_webtraffic}
\end{figure*}


%\bibliographystyle{elsarticle-num} 
%\bibliography{lib_new}

\end{document}